\pdfoutput=1
\documentclass[10pt,twocolumn,letterpaper]{article}

\usepackage[pagenumbers]{cvpr}

\usepackage{microtype}

\usepackage{multirow}
\usepackage{colortbl}
\usepackage{pifont}
\usepackage{bm}
\expandafter\def\csname ver@everyshi.sty\endcsname{2001/05/15 v3.00 EveryShipout Package (MS)}
\usepackage{tikz}
\usetikzlibrary{arrows.meta,positioning,calc,fit,backgrounds,patterns}
\usepackage{pgfplots}
\pgfplotsset{compat=1.16}

\definecolor{cOurs}{RGB}{0,114,178}
\definecolor{cRepa}{RGB}{230,159,0}
\definecolor{cSit}{RGB}{120,120,120}
\definecolor{cTok}{RGB}{0,158,115}
\definecolor{cZero}{RGB}{120,120,120}
\definecolor{cAlign}{RGB}{86,180,233}
\definecolor{cJoint}{RGB}{204,121,167} %
\definecolor{cLoc}{RGB}{213,94,0}   %

\usepackage{listings}
\usepackage{algorithm}
\lstdefinestyle{gcrepa}{basicstyle=\ttfamily\scriptsize, columns=fullflexible, keepspaces=true,
  morecomment=[l][\color{black!55}]{\#}, moredelim=**[is][\color{cOurs}]{@}{@},
  showstringspaces=false, aboveskip=2pt, belowskip=0pt}
\AtEndPreamble{\crefname{algorithm}{Alg.}{Algs.}\Crefname{algorithm}{Algorithm}{Algorithms}}

\newcommand{\ours}{GC-REPA\xspace}

\newcommand{\Mloc}{\ensuremath{M_{\text{within}}}\xspace}
\newcommand{\Mglob}{\ensuremath{M_{\text{image}}}\xspace}

\makeatletter
\renewcommand\LenToUnit[1]{\strip@pt\dimexpr(#1)*\p@/\unitlength\relax}
\makeatother

\definecolor{cvprblue}{rgb}{0.21,0.49,0.74}
\usepackage[pagebackref,breaklinks,colorlinks,allcolors=cvprblue]{hyperref}

\def\confName{CVPR}
\def\confYear{2027}

\title{Supply the Global, Supervise the Local:\\Combining Representation Conditioning and Alignment in Diffusion Transformers}

\author{Chaojie Ji\textsuperscript{1,2,$*$}\quad
Sergio Quiroga-Sandoval\textsuperscript{3,$*$}\quad
Ruxin Wang\textsuperscript{4,$\dagger$}\quad
Zheju Zheng\textsuperscript{4}\quad
Yankai Cao\textsuperscript{5,2,$\dagger$}\\[4pt]
{\normalsize\textsuperscript{1}Department of Mathematics, The University of British Columbia, Canada}\\
{\normalsize\textsuperscript{2}Institute of Applied Mathematics, The University of British Columbia, Canada}\\
{\normalsize\textsuperscript{3}Departamento de Matem\'aticas, Universidad Nacional de Colombia, Colombia}\\
{\normalsize\textsuperscript{4}Shenzhen Institutes of Advanced Technology, Chinese Academy of Sciences, Shenzhen, China}\\
{\normalsize\textsuperscript{5}Department of Chemical and Biological Engineering, The University of British Columbia, Canada}
}

\begin{document}
\maketitle
{\renewcommand{\thefootnote}{}\footnotetext{\textsuperscript{$*$}Equal contribution.\quad \textsuperscript{$\dagger$}Corresponding authors.}}
\begin{abstract}
A pretrained image representation helps train a diffusion transformer in several roles, including as an \emph{alignment target}, which the model's hidden states learn to match, as in representation alignment (REPA), and as a \emph{condition}, an extra input describing the image to be generated.
Combined, they would give the model an input and a target, but whether they conflict is unknown.
To test this, we let the condition carry only part of REPA's target: we split the representation of each image patch into an image-level (global) part, shared by all patches of the image, and a within-image (local) part, the patch's difference from the image-level part, and use one token carrying the image-level part as the condition.
Then, the two roles overlap on the image-level part, where they could conflict, while only REPA supervises the within-image part, where we can see whether the token helps REPA.

We find that the token and REPA \emph{together} teach both the image-level part and the within-image part, the main source of REPA's FID improvement, far more than the sum of what each teaches alone, so they cooperate rather than conflict.
These findings give one design rule for combining REPA with an image-level condition, \emph{supply the global, supervise the local}, which we implement as Global-Condition REPA (\mbox{GC-REPA}): REPA with the token as an extra input; the token carries no information beyond REPA's target and is drawn at sampling from a separately trained token prior.
\mbox{GC-REPA} greatly accelerates convergence: on ImageNet $256\times256$, SiT-XL/2 + \mbox{GC-REPA} reaches REPA's 4M- and SiT-XL/2's 7M-iteration FIDs within 200K and 100K iterations ($20\times$ and $70\times$ fewer).
\end{abstract}

\section{Introduction}
\label{sec:intro}

Diffusion models lead image generation~\cite{dhariwal2021adm,rombach2022ldm}, most recently as \emph{diffusion transformers}~\cite{peebles2023dit}, \eg, SiT (Scalable Interpolant Transformers)~\cite{ma2024sit}.
In these models, a stack of transformer blocks learns to recover an image's \emph{latent}, an image's compressed form from an autoencoder, from a \emph{noisy latent}, the latent mixed with noise~\cite{rombach2022ldm,albergo2023stochastic,lipman2023flow}.
They train slowly: SiT-XL/2 needs 7M iterations to reach its best Fr\'echet inception distance (FID)~\cite{ma2024sit} (\cref{fig:teaser}a).

\begin{figure}[t]
  \centering
  \begin{minipage}[b]{0.54\linewidth}
  \tikzset{
    speedup/.style={red!75!black, densely dashed, thick, -{Stealth[length=3.5pt]}},
    speedlab/.style={font=\tiny\bfseries, text=red!75!black, inner sep=1pt}}
  \begin{tikzpicture}
    \begin{axis}[xmode=log, log basis x=2, ymode=log, width=1.08\linewidth, height=4.6cm, scale only axis=false,
                 title={(a) Faster convergence (SiT-XL/2)}, title style={font=\scriptsize\bfseries, yshift=-1.2ex},
                 ylabel={FID $\downarrow$}, ylabel shift=-4pt, label style={font=\scriptsize}, tick label style={font=\tiny},
                 xmin=40, xmax=9500, xtick={100,400,1000,4000}, xticklabels={100K,400K,1M,4M},
                 ymin=2.5, ymax=90, ytick={3,5,10,20,40,80}, log ticks with fixed point,
                 grid=major, grid style={gray!20}, every axis plot/.append style={thick, mark size=1.3pt}]
      \addplot[cSit, mark=triangle*, densely dotted] coordinates {(50,78.9)(100,43.1)(150,32.0)(200,26.9)(400,17.2)(7000,8.3)};
      \addplot[cRepa, mark=square*, dashed] coordinates {(50,52.3)(100,19.4)(200,11.1)(400,7.9)(1000,6.4)(4000,5.9)};
      \addplot[cOurs, mark=*] coordinates {(50,19.3)(100,6.8)(200,4.0)(400,3.02)};
      \draw[speedup] (axis cs:7000,8.3) -- (axis cs:100,8.3);
      \node[speedlab, anchor=south] at (axis cs:900,8.3) {$70\times$};
      \draw[speedup] (axis cs:4000,5.9) -- (axis cs:200,5.9);
      \node[speedlab, anchor=north] at (axis cs:900,5.5) {$20\times$ faster};
      \node[font=\tiny, anchor=south, inner sep=1pt] at (axis cs:7000,8.6) {7M};
      \node[font=\tiny, text=cSit, anchor=south west, inner sep=1pt] at (axis cs:1300,12.5) {SiT};
      \node[font=\tiny, text=cRepa!85!black, anchor=west, inner sep=1pt] at (axis cs:112,21) {REPA};
      \node[font=\tiny, text=cOurs, anchor=west, inner sep=1pt] at (axis cs:52,3.0) {\ours};
    \end{axis}
  \end{tikzpicture}
  \end{minipage}\hfill
  \begin{minipage}[b]{0.44\linewidth}
  \resizebox{\linewidth}{!}{%
  \begin{tikzpicture}[font=\footnotesize,
      ptitle/.style={font=\footnotesize\bfseries, anchor=south west, inner sep=0pt},
      map/.style={inner sep=0pt, anchor=south west},
      mtitle/.style={font=\scriptsize\bfseries, align=center, inner sep=1.5pt}]
    \def\sz{1.4}
    \node[ptitle] at (0,4.25) {(b) Split of the representation};
    \node[map] at (0,2.3) {\includegraphics[width=\sz cm, height=\sz cm]{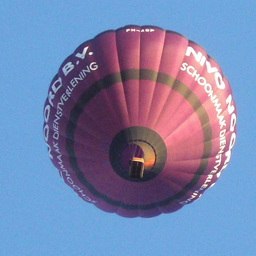}};
    \draw[gray!60, line width=0.3pt] (0,2.3) rectangle ++(\sz,\sz);
    \node[map] at (2.15,2.3) {\includegraphics[width=\sz cm, height=\sz cm]{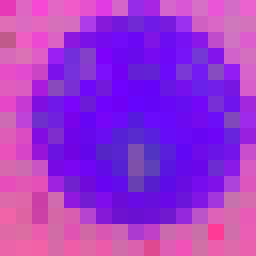}};
    \draw[gray!60, line width=0.3pt] (2.15,2.3) rectangle ++(\sz,\sz);
    \node[map] at (0,0) {\includegraphics[width=\sz cm, height=\sz cm]{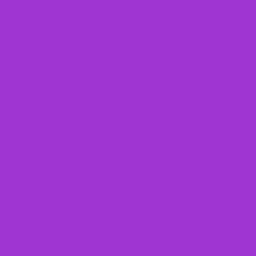}};
    \draw[gray!60, line width=0.3pt] (0,0) rectangle ++(\sz,\sz);
    \node[map] at (2.15,0) {\includegraphics[width=\sz cm, height=\sz cm]{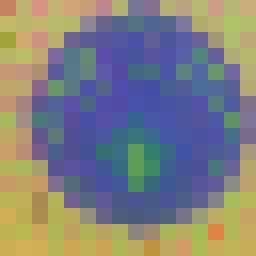}};
    \draw[gray!60, line width=0.3pt] (2.15,0) rectangle ++(\sz,\sz);
    \foreach \k in {1,...,15} {
      \draw[white, opacity=0.55, line width=0.15pt] ({\k*\sz/16},2.3) -- ++(0,\sz);
      \draw[white, opacity=0.55, line width=0.15pt] (0,{2.3+\k*\sz/16}) -- ++(\sz,0);
    }
    \foreach \k in {1,...,15} {
      \draw[white, opacity=0.35, line width=0.15pt] ({\k*\sz/16},0) -- ++(0,\sz);
      \draw[white, opacity=0.35, line width=0.15pt] (0,{\k*\sz/16}) -- ++(\sz,0);
    }
    \draw[-{Latex[length=1.1mm,width=1.2mm]}, line width=0.6pt, black!70] (1.47,3.0) -- (2.08,3.0)
      node[midway, above=1.5pt, font=\tiny, inner sep=0pt] {DINOv2};
    \node[mtitle, anchor=south] at (0.7,3.72) {image};
    \node[font=\scriptsize, anchor=north, inner sep=1pt, text=black!70] at (0.7,2.28) {$16\times16$ patches};
    \node[mtitle, anchor=south] at (2.85,3.72) {representation};
    \node[font=\scriptsize, anchor=north, align=center, inner sep=1pt, text=black!70] at (2.85,2.28) {256 vectors,\\one per patch};
    \node[font=\large, rotate=90] at (2.85,1.56) {$=$};
    \node[font=\large] at (1.775,0.7) {$+$};
    \node[mtitle, anchor=north, text=cTok!80!black] at (0.7,-0.02) {image-level part\\(the global)};
    \node[mtitle, anchor=north, text=cLoc] at (2.85,-0.02) {within-image part\\(the local)};
  \end{tikzpicture}}
  \end{minipage}
  \vspace{-4pt}
  \caption{\textbf{Faster convergence than SiT and REPA, and the split of the representation.}
  (a)~FID of SiT-XL/2 over training iterations (ImageNet $256\times256$); arrows: how many times fewer iterations \ours needs to reach the published FID of SiT at 7M and REPA at 4M iterations~\cite{ma2024sit,yu2025repa}.
  (b)~Each patch's representation is the image-level part, shared by all patches, plus the within-image part, the patch's difference from it (RGB colors: projection onto the top three principal components, the same in all maps; \cref{sec:visualization_protocol}).}
  \label{fig:teaser}
\end{figure}
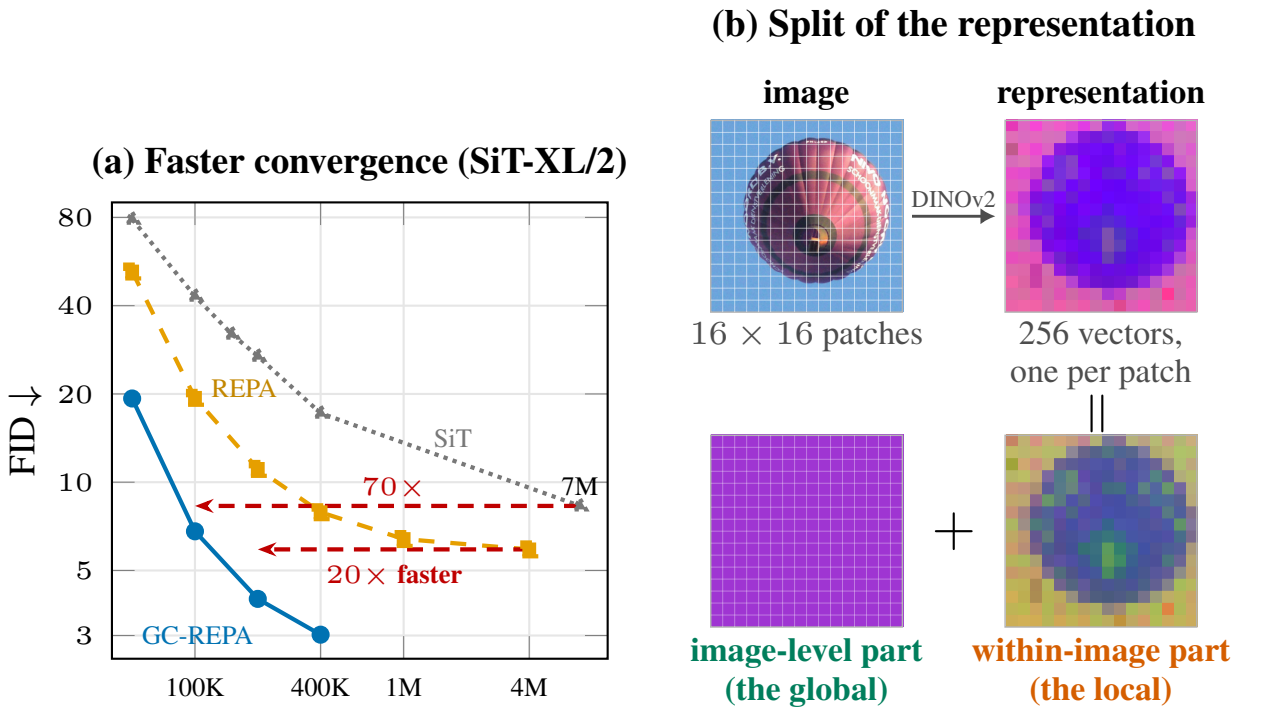

\begin{figure}[t]
  \centering
  \resizebox{\linewidth}{!}{%
  \begin{tikzpicture}[font=\footnotesize, >={Latex[length=1.3mm]},
      box/.style={draw, rounded corners=1.5pt, align=center, inner sep=1.5pt, minimum height=4.5mm, minimum width=5.5mm, fill=white},
      dit/.style={draw, rounded corners=2pt, fill=gray!8, minimum width=17mm, minimum height=8mm, align=center},
      ttl/.style={anchor=west, inner sep=0pt},
      note/.style={font=\scriptsize, inner sep=1pt, align=center}]
    \begin{scope}
      \node[ttl] at (-1.8,1.3) {\textbf{(a) Alignment target}};
      \node[note, anchor=north west, text=black!60] at (-1.8,1.08) {only in a loss};
      \node[dit] (d) at (1.6,-0.1) {diffusion\\transformer};
      \node[box] (z) at (-0.2,-0.1) {$\mathbf{z}_t$};
      \node[note] (e) at (-1.3,-0.1) {noise\\at level $t$};
      \draw[->, gray] (e) -- (z.west);
      \node[box] (o) at (3.0,-0.1) {$\hat{\mathbf{v}}$};
      \draw[->] (z) -- (d); \draw[->] (d) -- (o);
      \node[box, draw=cAlign!80!black, fill=cAlign!15] (y) at (1.6,1.0) {$\mathbf{y}_*$};
      \draw[<->, cAlign!70!black, thick, densely dashed] (y.south) -- (d.north) node[midway, right=3pt, note, text=cAlign!60!black] {REPA loss};
    \end{scope}
    \begin{scope}[xshift=5.45cm]
      \node[ttl] at (-1.8,1.3) {\textbf{(b) Condition}};
      \node[note, anchor=north west, text=black!60] at (-1.8,1.08) {received clean, not generated};
      \node[dit] (d) at (1.6,-0.1) {diffusion\\transformer};
      \node[box] (z) at (-0.2,-0.36) {$\mathbf{z}_t$};
      \node[box, draw=cTok!75!black, fill=cTok!15] (y) at (-0.2,0.17) {$\mathbf{y}_*$};
      \node[box] (o) at (3.0,-0.1) {$\hat{\mathbf{v}}$};
      \draw[->] (z) -- (z -| d.west);
      \draw[->, cTok!75!black, very thick] (y) -- node[above, note, text=cTok!60!black] {clean} (y -| d.west);
      \draw[->] (d) -- (o);
      \node[box, dashed, note, inner xsep=2pt, minimum height=5.5mm] (p) at (-1.3,0.13) {representa-\\tion prior};
      \draw[->, dashed] (p.east |- y) -- (y);
      \node[note, anchor=south, text=black!60] at (p.north) {(sampling)};
      \node[note] (e) at (-1.3,-0.50) {noise\\at level $t$};
      \draw[->, gray] (e) -- (z.west);
    \end{scope}
    \begin{scope}[yshift=-2.35cm]
      \node[ttl] at (-1.8,1.3) {\textbf{(c) Joint generation}};
      \node[dit] (d) at (1.6,-0.1) {diffusion\\transformer};
      \node[box] (z) at (-0.2,-0.36) {$\mathbf{z}_t$};
      \node[box, draw=cJoint!80!black, fill=cJoint!12, postaction={pattern=crosshatch dots, pattern color=cJoint!55}] (y) at (-0.2,0.17) {$\mathbf{y}_t$};
      \node[note] (e) at (-1.3,-0.1) {noise\\at level $t$};
      \draw[->, gray] (e) -- (y.west); \draw[->, gray] (e) -- (z.west);
      \draw[->] (z) -- (z -| d.west);
      \draw[->, cJoint!80!black, very thick] (y) -- node[above, note, text=cJoint!70!black] {noisy} (y -| d.west);
      \node[box] (o) at (3.0,-0.36) {$\hat{\mathbf{v}}$};
      \node[box, draw=cJoint!80!black, fill=cJoint!12] (oy) at (3.0,0.17) {$\hat{\mathbf{v}}_{\mathbf{y}}$};
      \draw[->] (o -| d.east) -- (o); \draw[->, cJoint!80!black, very thick] (oy -| d.east) -- (oy);
      \node[box, draw=cAlign!80!black, fill=cAlign!15] (ya) at (1.6,1.0) {$\mathbf{y}_*$};
      \draw[<->, cAlign!70!black, thick, densely dashed] (ya.south) -- (d.north) node[midway, right=3pt, note, text=cAlign!60!black] {REPA loss};
      \draw[->, cJoint!80!black, thick] (ya.west) -| node[pos=0.25, below, note, text=cJoint!70!black] {noising} (y.north);
    \end{scope}
    \begin{scope}[xshift=5.45cm, yshift=-2.35cm]
      \fill[cvprblue!7, rounded corners=3pt] (-1.95,-0.7) rectangle (3.4,1.5);
      \node[ttl] at (-1.8,1.3) {\textbf{(d) \ours}};
      \node[dit] (d) at (1.6,-0.1) {diffusion\\transformer};
      \node[box] (z) at (-0.2,-0.36) {$\mathbf{z}_t$};
      \node[box, draw=cTok!75!black, fill=cTok!15] (u) at (-0.2,0.17) {$\mathbf{g}$};
      \node[box] (o) at (3.0,-0.1) {$\hat{\mathbf{v}}$};
      \draw[->] (z) -- (z -| d.west);
      \draw[->, cTok!75!black, very thick] (u) -- node[above, note, text=cTok!60!black] {clean} (u -| d.west);
      \draw[->] (d) -- (o);
      \node[box, fill=gray!10] (y) at (1.6,1.0) {$\mathbf{y}_*$};
      \draw[<->, cAlign!70!black, thick, densely dashed] (y.south) -- (d.north) node[midway, right=3pt, note, text=cAlign!60!black] {REPA loss};
      \draw[->, cTok!75!black, thick] (y.west) -| node[pos=0.25, below, note, text=cTok!60!black] {patch average} (u.north);
      \node[box, dashed, note, inner xsep=2pt, minimum height=5.5mm] (p) at (-1.3,0.21) {token\\prior};
      \draw[->, dashed] (p.east |- u) -- (u);
      \node[note, anchor=south, text=black!60] at (p.north) {(sampling)};
      \node[note] (e) at (-1.3,-0.44) {noise\\at level $t$};
      \draw[->, gray] (e) -- (z.west);
    \end{scope}
  \end{tikzpicture}}
  \caption{\textbf{Roles of a pretrained representation} $\mathbf{y}_*$ (one vector per patch of the image) in training a diffusion transformer, which maps the noisy latent $\mathbf{z}_t$ to the predicted velocity $\hat{\mathbf{v}}$.
  (a)~Alignment target only. (b)~Condition only.
  (c)~Jointly generated output: noised and generated with the image ($\hat{\mathbf{v}}_{\mathbf{y}}$: its predicted velocity).
  (d)~\ours: the patch average $\mathbf{g}$ of $\mathbf{y}_*$ enters clean as the token, supplied at sampling by a token prior model; the REPA loss is unchanged.}
  \label{fig:roles}
\end{figure}
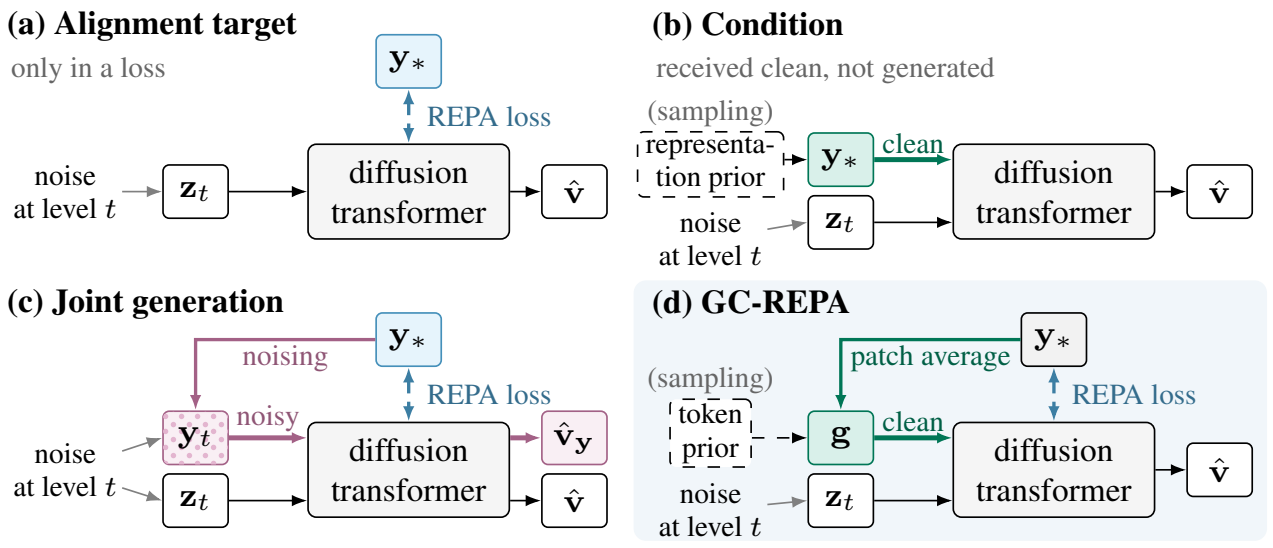

Representations from pretrained encoders, such as DINOv2~\cite{oquab2024dinov2}, which describes an image by one vector per \emph{patch}, the patch's representation (\cref{fig:teaser}b), make diffusion transformers train faster and generate better images, in three roles (\cref{fig:roles}a--c), each leaving a gap on its own.
As an \emph{alignment target}, the representation appears only in a training loss, as in REPA (representation alignment)~\cite{yu2025repa}: the \emph{REPA loss} pulls the hidden states of a transformer block toward the representation and is added to the standard denoising loss of transformer blocks.
The representation never reaches the diffusion transformer as input, so the hidden states must recover it from the noisy latent alone, and at high noise, where this latent shows little of the image, this recovery is poor.
As a \emph{condition}, the representation is an extra input that tells the diffusion transformer about the image to be generated; in RCG~\cite{li2024rcg}, at image sampling, where there is no image to provide the condition, a separate model supplies it.
A condition gives the representation directly but sets no target for the hidden states.
As a \emph{jointly generated output}, the representation is instead noised with the latent and denoised together with it~\cite{wu2025reg,kouzelis2025redi}, optionally with the REPA loss; so at high noise it shows as little as the noisy latent does.

The natural alternative, using the same representation both as a clean condition and as an alignment target, here REPA, would close all three gaps, but the two roles may conflict. REPA teaches the hidden states to reproduce the representation, which the condition may already supply, so REPA could make the condition redundant or work against it.

To see how the two roles interact, we must know where they overlap and where they do not: where both carry the same information, they may conflict; where only REPA supervises, we can see whether the condition helps REPA.
We therefore split the patch's representation, itself REPA's target, into two parts and let the condition, a single \emph{token}, take only the image-level part, since a condition typically describes the image as a whole.
Concretely, the \emph{image-level part} (\cref{fig:teaser}b) is the \emph{patch average} of the representation, the one vector closest to all patches of the image, which describes what the image shows as a whole; the \emph{within-image part} is the patch's difference from the patch average, which describes how the patch differs from the image as a whole.
The two roles then overlap on the image-level part, where we test whether they conflict, and only REPA supervises the within-image part, where we test whether the token helps REPA.

The split also makes the roles' interaction measurable part by part: since both roles act on the hidden states, we measure how much of each part the hidden states hold and which part improves image quality. We find: together, the token and REPA teach the image-level part far more than the sum of what each teaches alone (they do not conflict: Finding~1), and the token, mainly if clean and from the image itself, helps REPA teach the within-image part, although the token alone barely improves it (they do cooperate: Finding~2); and REPA improves FID mainly through the within-image part (Finding~3).

These findings give one design rule for combining REPA with an image-level condition, \emph{supply the global, supervise the local}: supply the image-level part (the global) as a condition (Finding~1), and teach the within-image part (the local) with REPA (Findings~2 and~3).
We implement this rule as Global-Condition REPA (\ours, \cref{fig:roles}d): the patch-average token, drawn at sampling from a separate token prior, supplies the global; the unchanged REPA loss supervises the local by its within-image term (Finding~3), and with the token, its image-level term does no harm.

\ours is \emph{simple}, and what it adds is \emph{small}: its token, derived from REPA's own target, uses no information beyond it, and it changes only a few lines of REPA's training code; it adds a token (0.6--0.9M parameters) and a separately trained token prior, which adds at most 0.2\% to the sampling compute.
It is also \emph{effective}: at 400K iterations, it lowers REPA's FID from 24.4 to 9.3 on SiT-B/2 and from 10.0 to 3.5 on SiT-L/2~\cite{ma2024sit}, and on SiT-XL/2 it reaches REPA's 4M-iteration FID within 200K iterations ($20\times$ fewer, \cref{fig:teaser}a).

In summary, our contributions are:
\begin{itemize}[leftmargin=*,itemsep=1pt,topsep=2pt]
\item Splitting the representation into image-level and within-image parts, we show that a token carrying the image-level part and REPA cooperate rather than conflict in teaching both parts (Findings~1 and~2).
\item Aligning each part separately, we find REPA's generation gain mainly in the within-image part (Finding~3), completing the rule \emph{supply the global, supervise the local}, which \ours implements by adding to REPA a token and a separately trained token prior.
\item \ours is simple and adds little computation yet greatly speeds up training: it reaches REPA's and SiT's published ImageNet FIDs with $20\times$ and $70\times$ fewer iterations.
\end{itemize}
\section{Preliminaries}
\label{sec:prelim}

\label{sec:prelim_bg}
\paragraph{Flow-based diffusion transformers.}
We follow SiT~\cite{ma2024sit}, which generates images in the compressed latent space of a pretrained variational autoencoder (VAE) within the stochastic interpolant framework~\cite{albergo2023stochastic,lipman2023flow,liu2023rectified}.
The VAE's encoder $E$~\cite{rombach2022ldm} maps an image $\mathbf{x}$ to its latent $\mathbf{z}_* = E(\mathbf{x})$.
The path from this latent to noise is, with $\boldsymbol{\epsilon}\sim\mathcal{N}(\mathbf{0},\mathbf{I})$, $\mathbf{z}_t = \alpha_t\mathbf{z}_* + \sigma_t\boldsymbol{\epsilon}$ at noise level $t\in[0,1]$, with $\alpha_t = 1-t$ and $\sigma_t = t$, so $t=0$ gives the original latent and $t=1$ pure noise.
A transformer $\mathbf{v}_\theta$ learns the \emph{velocity} of this path, its derivative w.r.t. $t$, $\mathbf{v}^\star_t = \dot{\alpha}_t\mathbf{z}_* + \dot{\sigma}_t\boldsymbol{\epsilon}$.
Specifically, the transformer $\mathbf{v}_\theta$ cuts $\mathbf{z}_t$ into $N$ latent patches, embeds each, and passes the $N$ embeddings through a stack of transformer blocks, each of which outputs one vector per latent patch; conditioned on $t$ and the class label $c$, it minimizes the \emph{denoising loss}
\begin{equation}
  \mathcal{L}_{\text{vel}}(\theta) =
  \mathbb{E}_{\mathbf{z}_*,\boldsymbol{\epsilon},t}\big[\,\|\mathbf{v}_\theta(\mathbf{z}_t,t,c)-\mathbf{v}^\star_t\|_2^2\,\big].
  \label{eq:vel}
\end{equation}
Sampling follows the path backward: it integrates the reverse-time stochastic differential equation (SDE) from $t=1$ to $t=0$ with SiT's Euler--Maruyama sampler.

\paragraph{Representation alignment.}
Let $\mathbf{y}_* = f(\mathbf{x})\in\mathbb{R}^{N\times D}$ be the \emph{representation} of the image under a frozen pretrained encoder $f$, \eg DINOv2: one $D$-dimensional vector $\mathbf{y}_*^{[n]}$ for each of $N$ image patches, aligned one to one with the diffusion transformer's latent patches; below, patch $n$ refers to both.
Let $\mathbf{h}_t\in\mathbb{R}^{N\times D'}$ be the hidden states that transformer block $\ell$ of $\mathbf{v}_\theta$ outputs for $\mathbf{z}_t$ (we drop $\ell$ from the notation), one vector $\mathbf{h}_t^{[n]}$ per patch, with $D'$ the hidden size.
REPA~\cite{yu2025repa} maps the hidden states with a trainable MLP projector $h_\phi$ and maximizes their patch-wise cosine similarity $\mathrm{sim}(\mathbf{a},\mathbf{b})$ to the representation,
\begin{equation}
  \mathcal{L}_{\text{REPA}}(\theta,\phi) =
  -\mathbb{E}_{\mathbf{x},\boldsymbol{\epsilon},t}\Big[\frac{1}{N}\sum_{n=1}^{N}
  \mathrm{sim}\big(\mathbf{y}_*^{[n]},\,h_\phi(\mathbf{h}_t^{[n]})\big)\Big],
  \label{eq:repa}
\end{equation}
training with $\mathcal{L}_{\text{vel}}+\lambda\mathcal{L}_{\text{REPA}}$. We call $\mathcal{L}_{\text{REPA}}$ the \emph{REPA loss}.
We keep REPA's defaults: $\lambda=0.5$, and $\ell=4$ for SiT-B/2 and $\ell=8$ for SiT-L/2 and SiT-XL/2.
Only blocks $1,\dots,\ell$, \ie, the \emph{aligned blocks}, receive REPA's gradient.

\section{Two Roles, Two Parts}
\label{sec:analysis}

The token, the input of \ours (\cref{fig:roles}d), carries only the image-level part, which REPA also supervises, so on this part REPA could make the token redundant or work against it; on the within-image part, which only REPA supervises, the token could instead help REPA (\cref{sec:design}).
\Cref{sec:supply,sec:placement} test these two possibilities, and \cref{sec:supervise} tests which part gives REPA's FID gain.
\subsection{Study design}
\label{sec:design}
\paragraph{Split of the representation.}
Intuitively, the representations of all patches of an image share what the image shows as a whole (in \cref{fig:teaser}b, a balloon in the sky), and each patch's representation differs in what the patch shows locally (fabric or burner in \cref{fig:teaser}b).
Averaging over the patches keeps what they share and cancels what differs, so the image-level part is the average of the patches' representations, the single vector closest to all of them; the within-image part is each patch's deviation from that average, what the patch shows locally.
In symbols, with the unit-normalized patch representations $\hat{\mathbf{y}}^{[n]}=\mathbf{y}_*^{[n]}/\|\mathbf{y}_*^{[n]}\|_2$, their image mean $\bar{\mathbf{y}}=\frac{1}{N}\sum_{n}\hat{\mathbf{y}}^{[n]}$, and each patch's difference from that mean, $\tilde{\mathbf{y}}^{[n]} = \hat{\mathbf{y}}^{[n]}-\bar{\mathbf{y}}$, each patch splits as
$\hat{\mathbf{y}}^{[n]} = \underbrace{\bar{\mathbf{y}}}_{\text{image-level part}} + \underbrace{\tilde{\mathbf{y}}^{[n]}}_{\text{within-image part}}$.

REPA's loss compares each patch's representation $\mathbf{y}_*^{[n]}$ with the patch's projected hidden state $h_\phi(\mathbf{h}_t^{[n]})$ (\cref{eq:repa}); to see how this loss acts on each part, we split the projected hidden states in the same way: with $\hat{\mathbf{h}}^{[n]}$ the unit-normalized $h_\phi(\mathbf{h}_t^{[n]})$, $\bar{\mathbf{h}}$ its image mean and $\tilde{\mathbf{h}}^{[n]} = \hat{\mathbf{h}}^{[n]}-\bar{\mathbf{h}}$, the similarity in REPA's loss splits as
\begin{equation}
  \begin{aligned}
  &\frac{1}{N}\sum_{n=1}^{N}\mathrm{sim}\big(\mathbf{y}_*^{[n]},h_\phi(\mathbf{h}_t^{[n]})\big)
  = \frac{1}{N}\sum_{n=1}^{N}\hat{\mathbf{y}}^{[n]\top}\hat{\mathbf{h}}^{[n]}\\
  &\qquad= \underbrace{\bar{\mathbf{y}}^{\top}\bar{\mathbf{h}}}_{\text{image-level}}
  + \underbrace{\frac{1}{N}\sum_{n=1}^{N}\tilde{\mathbf{y}}^{[n]\top}\tilde{\mathbf{h}}^{[n]}}_{\text{within-image}},
  \end{aligned}
  \label{eq:shares}
\end{equation}
because $\sum_{n}\tilde{\mathbf{y}}^{[n]} = \sum_{n}\tilde{\mathbf{h}}^{[n]} = \mathbf{0}$ (\cref{sec:proof_shares}), which holds only when the image-level part is the patch average.
The split lets us measure how well the hidden states of transformer blocks match each part and align one part at a time.

The token and REPA overlap, and so could conflict, only through the image-level term of \cref{eq:shares}, and this term is not small: at perfect alignment, where the similarity is 1, the image-level term $\|\bar{\mathbf{y}}\|^2$ gives 0.29 for DINOv2-B on ImageNet and 0.42 for CLIP-L~\cite{radford2021clip}.

\paragraph{The token: the condition we study.}
\label{sec:token}
Our condition, the \emph{token}, carries the image-level part: the \emph{patch average} $\mathbf{g}$, \ie, $\bar{\mathbf{y}}$ rescaled to length 1, embedded as $\mathbf{e}$:
\begin{equation}
  \mathbf{g} = \frac{\bar{\mathbf{y}}}{\|\bar{\mathbf{y}}\|_2},
  \qquad
  \mathbf{e} = \mathrm{LN}\big(\mathbf{A}\,s(\mathbf{g})+\mathbf{b}\big)\in\mathbb{R}^{D'},
  \label{eq:token}
\end{equation}
where $s(\cdot)$ standardizes $\mathbf{g}$ over its $D$ channels, and a linear layer ($\mathbf{A}$, $\mathbf{b}$) and LayerNorm ($\mathrm{LN}$)~\cite{ba2016layernorm} map $\mathbf{g}$ to the hidden size $D'$.
Then, transformer blocks take $\mathbf{e}$ as one more input next to the $N$ patch embeddings; in each block, the patches can draw on $\mathbf{e}$ through attention, just as they draw on one another; the token is only an input, and its output is discarded. During training, the token is dropped (\ie, $s(\mathbf{g})$ is set to zero) for a random 20\% of the samples, so the diffusion transformer also learns to run without it.
At sampling, where there is no image, $\mathbf{g}$ is drawn from the \emph{token prior}, a class-conditional model of $\mathbf{g}$ (\cref{sec:prior}).

To test whether the interaction between token and REPA depends on where the token enters w.r.t. the aligned blocks, we place the token in two ways: the \emph{early token} enters before block~1, so every block can read it, and the \emph{late token} enters after block~$\ell$, so only the blocks after the aligned blocks can.

\begin{table}[t]
  \centering
  \small
  \setlength{\tabcolsep}{3pt}
  \newcommand{\sd}[1]{{\scriptsize$\pm$#1}}
  \resizebox{\linewidth}{!}{%
  \begin{tabular}{@{}llcccc@{}}
    \toprule
    Token & REPA & \Mglob$\uparrow$ & \Mloc$\uparrow$ & FID$\downarrow$ & sFID$\downarrow$ \\
    \midrule
    -- & -- & 0.260\sd{0.001} & 0.315\sd{0.001} & 93.97\sd{1.08} & 14.19\sd{0.32} \\
    -- & full & 0.319\sd{0.002} & 0.325\sd{0.000} & 77.66\sd{0.36} & 15.61\sd{1.55} \\
    early & -- & 0.539\sd{0.005} & 0.330\sd{0.002} & 67.41\sd{1.02} & \phantom{0}9.32\sd{0.28} \\
    early & full & 0.836\sd{0.012} & 0.399\sd{0.001} & 47.03\sd{0.10} & \phantom{0}8.06\sd{0.17} \\
    \midrule
    late & -- & 0.245\sd{0.003} & 0.316\sd{0.001} & 70.97\sd{0.84} & \phantom{0}9.70\sd{0.59} \\
    late & full & 0.318\sd{0.001} & 0.326\sd{0.000} & 61.20\sd{0.26} & \phantom{0}9.06\sd{0.06} \\
    \midrule
    -- & image-level & 0.295\sd{0.003} & 0.315\sd{0.000} & 84.27\sd{0.40} & 15.87\sd{0.53} \\
    -- & within-image & 0.326\sd{0.002} & 0.327\sd{0.001} & 73.89\sd{0.79} & 14.75\sd{2.48} \\
    early & image-level & 0.939\sd{0.002} & 0.319\sd{0.003} & 65.97\sd{0.73} & \phantom{0}8.89\sd{0.23} \\
    early & within-image & 0.627\sd{0.003} & 0.379\sd{0.001} & 47.55\sd{0.67} & \phantom{0}8.54\sd{0.24} \\
    \bottomrule
  \end{tabular}}
  \caption{\textbf{Token placement $\times$ REPA target} (SiT-B/2, 50K iterations; mean $\pm$ std.).
  \Mglob, \Mloc: linear-probe scores for the image-level and within-image parts.
  Token: none (--), \emph{early} or \emph{late} (before or after the aligned blocks).
  REPA: none (--), full, or one part only; more metrics: \cref{tab:supp_interaction}.}
  \label{tab:interaction}\label{tab:target}
\end{table}

\paragraph{Common setup and scores.}
The three findings below compare models that differ only in the token and in what REPA aligns, each comparison answering one of the three tests named at the start of this section.
First, we train SiT-B/2 with DINOv2-B on ImageNet from scratch for 50K iterations with REPA's recipe, using three seeds, and report the mean and standard deviation (\cref{tab:interaction}); SiT-L/2 and CLIP-L give similar results (\cref{sec:supp_variants}).
We then measure how much of each part (image-level or within-image) the hidden states of the last aligned block hold, with two scores, the \emph{cosine} and the \emph{linear probe}, which measure the similarity between the part and these hidden states with or without REPA's projector. We say that the token or REPA \emph{teaches} a part when it raises these scores.
Both scores are computed at every noise level; unless stated otherwise, we report them at $t=0.75$, where the noisy latent shows little of the image-level part (results at all $t$: \cref{fig:mechanism,fig:supp_probe_t}).
With longer training (400K iterations), both scores of all models change, but their ranking of the models is unchanged.
Additionally, we measure sample quality by FID~\cite{heusel2017fid} and sFID~\cite{nash2021sfid} on 50K samples; other metrics and details are in \cref{sec:supp_full,sec:supp_impl}.

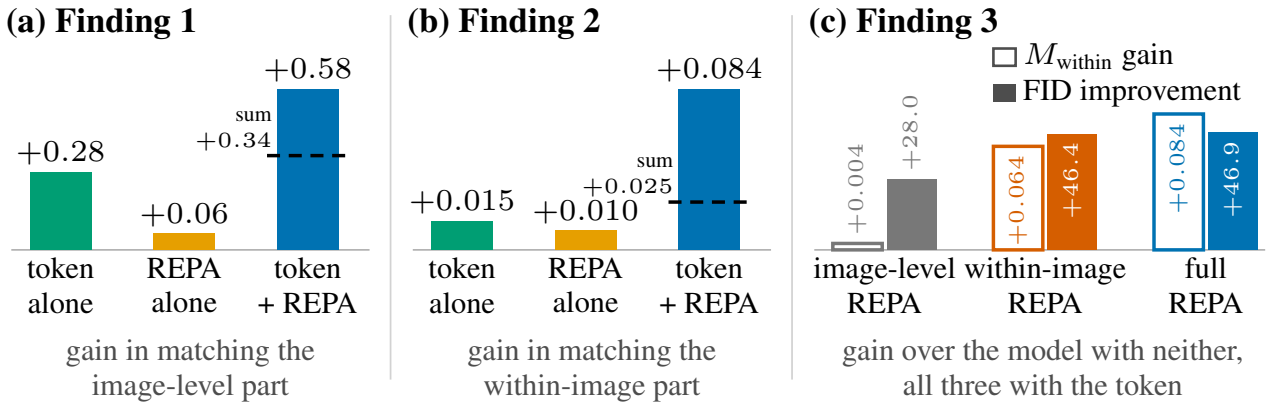
\begin{figure}[t]
  \centering
  \resizebox{\linewidth}{!}{%
  \begin{tikzpicture}[font=\footnotesize,
      ptitle/.style={font=\footnotesize\bfseries, anchor=south west, inner sep=0pt},
      val/.style={font=\scriptsize, anchor=south, inner sep=1.5pt},
      cat/.style={font=\scriptsize, anchor=north, align=center, inner sep=1.5pt},
      metric/.style={font=\scriptsize, anchor=north, align=center, text=black!70}]
    \foreach \x in {3.12,6.37} {\draw[gray!35] (\x,-1.05) -- (\x,1.9);}
    \begin{scope}[shift={(0,0)}]
      \node[ptitle] at (0,1.7) {(a) Finding 1};
      \draw[gray!60] (0.05,0) -- (2.95,0);
      \fill[cTok]   (0.2,0) rectangle (0.7,0.631);  \node[val] at (0.45,0.631) {$+0.28$};
      \fill[cRepa] (1.2,0) rectangle (1.7,0.132);  \node[val] at (1.45,0.132) {$+0.06$};
      \fill[cOurs]  (2.2,0) rectangle (2.7,1.3);   \node[val] at (2.45,1.3) {$+0.58$};
      \draw[black, densely dashed, line width=0.8pt] (2.12,0.763) -- (2.78,0.763);
      \node[font=\tiny, anchor=south east, align=right, inner sep=1pt] at (2.2,0.763) {sum\\$+0.34$};
      \node[cat] at (0.45,0) {token\\alone};
      \node[cat] at (1.45,0) {REPA\\alone};
      \node[cat] at (2.45,0) {token\\+ REPA};
      \node[metric] at (1.5,-0.6) {gain in matching the\\image-level part};
    \end{scope}
    \begin{scope}[shift={(3.25,0)}]
      \node[ptitle] at (0,1.7) {(b) Finding 2};
      \draw[gray!60] (0.05,0) -- (2.95,0);
      \fill[cTok]   (0.2,0) rectangle (0.7,0.232);  \node[val] at (0.45,0.232) {$+0.015$};
      \fill[cRepa] (1.2,0) rectangle (1.7,0.155);  \node[val] at (1.45,0.155) {$+0.010$};
      \fill[cOurs]  (2.2,0) rectangle (2.7,1.3);   \node[val] at (2.45,1.3) {$+0.084$};
      \draw[black, densely dashed, line width=0.8pt] (2.12,0.387) -- (2.78,0.387);
      \node[font=\tiny, anchor=south east, align=right, inner sep=1pt] at (2.2,0.387) {sum\\$+0.025$};
      \node[cat] at (0.45,0) {token\\alone};
      \node[cat] at (1.45,0) {REPA\\alone};
      \node[cat] at (2.45,0) {token\\+ REPA};
      \node[metric] at (1.5,-0.6) {gain in matching the\\within-image part};
    \end{scope}
    \begin{scope}[shift={(6.5,0)}]
      \node[ptitle] at (0,1.7) {(c) Finding 3};
      \draw[gray!60] (0.05,0) -- (3.70,0);
      \draw[black!70, line width=0.6pt] (1.50,1.49) rectangle (1.68,1.63); \node[font=\scriptsize, anchor=west, inner sep=1pt] at (1.70,1.56) {\Mloc gain};
      \fill[black!70] (1.50,1.20) rectangle (1.68,1.34); \node[font=\scriptsize, anchor=west, inner sep=1pt] at (1.70,1.27) {FID improvement};
      \draw[cZero, line width=0.8pt] (0.20,0) rectangle (0.60,0.052);
      \node[font=\tiny, text=cZero, rotate=90, anchor=west, inner sep=1pt] at (0.40,0.08) {$+0.004$};
      \fill[cZero] (0.64,0) rectangle (1.04,0.567);
      \node[font=\tiny, text=cZero, rotate=90, anchor=west, inner sep=1pt] at (0.84,0.595) {$+28.0$};
      \draw[cLoc, line width=0.8pt]  (1.50,0) rectangle (1.90,0.838);
      \node[font=\tiny, text=cLoc, rotate=90, anchor=east, inner sep=1pt] at (1.70,0.81) {$+0.064$};
      \fill[cLoc]  (1.94,0) rectangle (2.34,0.939);
      \node[font=\tiny, text=white, rotate=90, anchor=east, inner sep=1pt] at (2.14,0.91) {$+46.4$};
      \draw[cOurs, line width=0.8pt] (2.80,0) rectangle (3.20,1.1);
      \node[font=\tiny, text=cOurs, rotate=90, anchor=east, inner sep=1pt] at (3.00,1.07) {$+0.084$};
      \fill[cOurs] (3.24,0) rectangle (3.64,0.95);
      \node[font=\tiny, text=white, rotate=90, anchor=east, inner sep=1pt] at (3.44,0.92) {$+46.9$};
      \node[cat] at (0.62,0) {image-level\\REPA};
      \node[cat] at (1.92,0) {within-image\\REPA};
      \node[cat] at (3.22,0) {full\\REPA};
      \node[metric] at (1.9,-0.6) {gain over the model with neither,\\all three with the token};
    \end{scope}
  \end{tikzpicture}}
  \caption{\textbf{The three findings} (SiT-B/2; higher is better; each panel and metric on its own scale).
  (a,\,b)~Gain in \Mglob (a) and \Mloc (b) over the model with neither token nor REPA; dashed: the sum of the two separate gains.
  (c)~Gain in \Mloc (hollow) and FID improvement (filled), over the model with neither token nor REPA, of the token with REPA aligning only the image-level part, only the within-image part, or both (full).}
  \label{fig:findings}
\end{figure}

\subsection{Finding 1: the image-level part}
\label{sec:supply}

The token and REPA together teach the image-level part far more than the sum of what each teaches alone (\cref{fig:findings}a). We show this with the two scores and explain it by how much the aligned blocks attend to the token.

\paragraph{Setup: measuring the image-level part.}
We measure how well the hidden states of the last aligned block match the image-level part $\bar{\mathbf{y}}$ by two scores.
The \emph{image-level cosine} measures this match through REPA's projector.
Since only models trained with REPA have the projector, the \emph{image-level probe} \Mglob replaces the projector with a fitted linear map, so that models without REPA can be measured too.

These two scores both use the \emph{centered cosine} $\mathrm{csim}(\mathbf{a},\bar{\mathbf{y}}) = \mathrm{sim}(\mathbf{a}-\mathbb{E}[\mathbf{a}],\,\bar{\mathbf{y}}-\mathbb{E}[\bar{\mathbf{y}}])$, with $\mathbb{E}$ the mean over images; centering removes what all images share, so the score reflects how images differ.
The image-level cosine sets $\mathbf{a}$ to $\bar{\mathbf{h}}$, the image-level part of the projected hidden states (\cref{eq:shares}).
For \Mglob, we fit a linear map $(\mathbf{V},\mathbf{u})$ from the image mean of the hidden states, $\mathbf{m}_t=\frac{1}{N}\sum_n\mathbf{h}_t^{[n]}$, to $\bar{\mathbf{y}}$. \Mglob is then the mean of $\mathrm{csim}\big(\mathbf{V}\mathbf{m}_t+\mathbf{u},\,\bar{\mathbf{y}}\big)$ over images not used for fitting (more details in \cref{sec:supp_probes}).

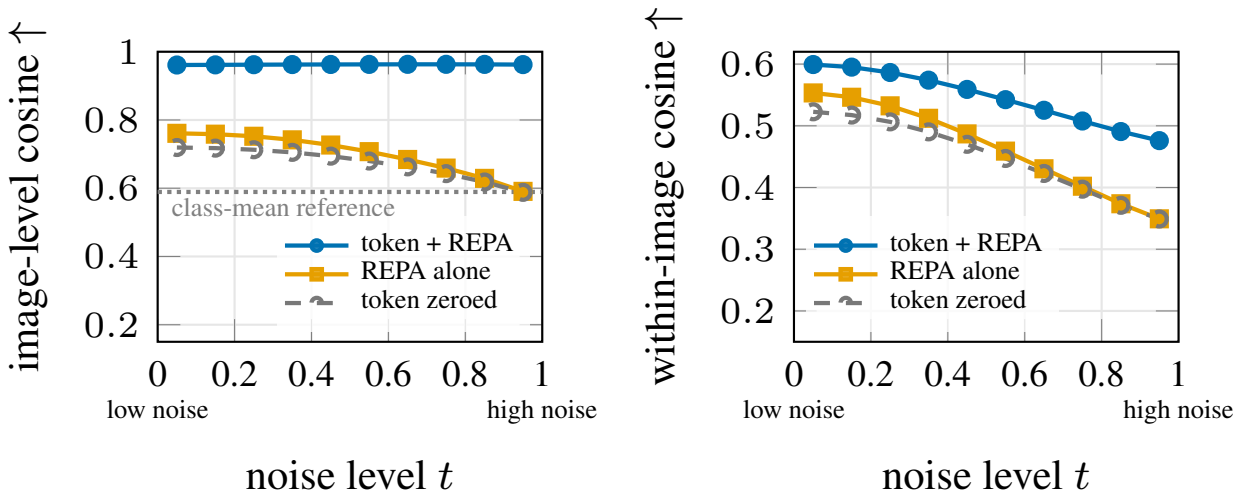
\begin{figure}[t]
  \centering
  \pgfplotsset{
    mech/.style={
      width=\linewidth, height=3.5cm,
      tick label style={font=\scriptsize}, label style={font=\footnotesize},
      legend style={font=\tiny, draw=none, fill=white, fill opacity=0.8, text opacity=1, row sep=-3.5pt, inner sep=1pt},
      legend image post style={scale=0.7},
      legend cell align=left, grid=major, grid style={gray!20},
      every axis plot/.append style={thick, mark size=1.4pt},
      xtick={0,0.2,0.4,0.6,0.8,1},
      xticklabels={{0\\[-2pt]{\tiny low noise}},0.2,0.4,0.6,0.8,{1\\[-2pt]{\tiny high noise}}},
      xticklabel style={align=center},
    }}
  \begin{subfigure}[t]{0.495\linewidth}
    \centering
    \begin{tikzpicture}
      \begin{axis}[mech, xlabel={noise level $t$}, ylabel={image-level cosine $\uparrow$},
                   xmin=0, xmax=1, ymin=0.15, ymax=1.0, legend style={at={(0.97,0.03)}, anchor=south east}]
        \addplot[cOurs, mark=*] coordinates {(0.05,0.9611)(0.15,0.9615)(0.25,0.9620)(0.35,0.9624)(0.45,0.9627)(0.55,0.9629)(0.65,0.9631)(0.75,0.9631)(0.85,0.9629)(0.95,0.9623)};
        \addplot[cRepa, mark=square*] coordinates {(0.05,0.7609)(0.15,0.7584)(0.25,0.7523)(0.35,0.7418)(0.45,0.7266)(0.55,0.7072)(0.65,0.6843)(0.75,0.6590)(0.85,0.6289)(0.95,0.5908)};
        \addplot[cZero, dashed, mark=o] coordinates {(0.05,0.7193)(0.15,0.7175)(0.25,0.7128)(0.35,0.7051)(0.45,0.6940)(0.55,0.6797)(0.65,0.6626)(0.75,0.6428)(0.85,0.6182)(0.95,0.5879)};
        \addplot[gray, densely dotted, mark=none, domain=0:1, forget plot] {0.5891};
        \node[font=\tiny, text=gray, anchor=north west, inner sep=1pt] at (axis cs:0.02,0.5891) {class-mean reference};
        \legend{token + REPA, REPA alone, token zeroed}
      \end{axis}
    \end{tikzpicture}
    \caption{Image-level part.}
    \label{fig:mech_global}
  \end{subfigure}\hfill
  \begin{subfigure}[t]{0.495\linewidth}
    \centering
    \begin{tikzpicture}
      \begin{axis}[mech, xlabel={noise level $t$}, ylabel={within-image cosine $\uparrow$},
                   xmin=0, xmax=1, ymin=0.15, ymax=0.62, ytick={0.2,0.3,...,0.6}, legend pos=south west]
        \addplot[cOurs, mark=*] coordinates {(0.05,0.5993)(0.15,0.5951)(0.25,0.5864)(0.35,0.5740)(0.45,0.5590)(0.55,0.5424)(0.65,0.5252)(0.75,0.5078)(0.85,0.4907)(0.95,0.4760)};
        \addplot[cRepa, mark=square*] coordinates {(0.05,0.5533)(0.15,0.5463)(0.25,0.5324)(0.35,0.5122)(0.45,0.4869)(0.55,0.4590)(0.65,0.4304)(0.75,0.4021)(0.85,0.3737)(0.95,0.3493)};
        \addplot[cZero, dashed, mark=o] coordinates {(0.05,0.5227)(0.15,0.5174)(0.25,0.5064)(0.35,0.4902)(0.45,0.4700)(0.55,0.4471)(0.65,0.4228)(0.75,0.3973)(0.85,0.3712)(0.95,0.3490)};
        \legend{token + REPA, REPA alone, token zeroed}
      \end{axis}
    \end{tikzpicture}
    \caption{Within-image part.}
    \label{fig:mech_local}
  \end{subfigure}
  \caption{\textbf{Both parts across noise levels, with and without the token} (SiT-B/2, 50K): image-level and within-image cosine at block~4.
  Dotted in (a): the \emph{class-mean reference} (hidden states replaced by the class mean of $\bar{\mathbf{y}}$); \emph{token zeroed}: the trained token + REPA model, fed a zeroed token.}
  \label{fig:mechanism}
\end{figure}

\paragraph{Observations.}
\emph{1. REPA alone teaches the image-level part only partly.}
REPA alone raises \Mglob only slightly over a model with neither the token nor REPA (\cref{tab:interaction}), and its image-level cosine drops as the noise grows (\cref{fig:mech_global}).
At the highest noise, REPA alone falls to the \emph{class-mean reference} (\cref{sec:supp_input}), the image-level cosine obtained by replacing the hidden states with the class mean of $\bar{\mathbf{y}}$, \ie, seeing only the class (dotted in \cref{fig:mech_global}).
So at high noise, the noisy latent reveals little of the image-level part beyond the class, limiting how much of this part REPA can teach.

\emph{2. The token alone supplies more of it, but not all.}
The token alone raises \Mglob much more (\cref{tab:interaction}), as the token supplies the image-level part directly.
But the aligned blocks use the token only moderately: %
zeroing the token changes each aligned block's hidden states by only 18\% (\cref{fig:mech_usage}).

\emph{3. Together, the token and REPA far exceed the sum of their separate gains.}
The token and REPA together raise \Mglob by 0.58, \emph{against 0.34 for the sum of their separate gains}, and their image-level cosine stays at 0.96 at every noise level (\cref{fig:mech_global}).
With REPA, the aligned blocks use the token much more: block~1 sends 67\% of its attention to the token, against 31\% with the token alone, and zeroing the token changes the aligned blocks about three times as much (\cref{sec:atten-exp-settings}), so the two roles do not conflict.

\subsection{Finding 2: the within-image part}
\label{sec:placement}

With the token, REPA teaches the within-image part better than REPA alone. We show this with the scores and explore what the gain needs: a clean token carrying the image-level part of the image itself, entering before the aligned blocks.

\paragraph{Setup: measuring the within-image part.}
The two scores mirror those used in Finding~1, with the within-image parts in place of the image-level ones and the plain cosine in place of the centered one. Specifically, the \emph{within-image cosine} compares $\tilde{\mathbf{h}}^{[n]}$ with $\tilde{\mathbf{y}}^{[n]}$, and the \emph{within-image probe} \Mloc fits a linear map from $\mathbf{h}_t^{[n]}-\mathbf{m}_t$ to $\tilde{\mathbf{y}}^{[n]}$ and averages the cosine of its predictions over the patches of images not used for fitting.
Since the image mean is removed from both sides, the image-level part does not enter either score.

\paragraph{Observations.}
\emph{1. Alone, the token barely improves the within-image part, and REPA does so only at low noise.}
The token alone raises \Mloc by at most 0.016 at every noise level; REPA alone raises it at low noise, where the noisy latent still shows the image, but barely at higher noise ($t\ge0.75$; \cref{tab:interaction,fig:supp_probe_t}).

\emph{2. REPA and the token together add far more than the sum of their separate gains in the within-image part, most at high noise.}
Together they raise \Mloc by \emph{3.4 times the sum of their separate gains} (\cref{fig:findings}b), so the token helps REPA.
With the token, REPA's within-image cosine exceeds that of REPA alone by 0.13 at the highest noise, against 0.05 at the lowest (\cref{fig:mech_local}), so the gain is largest where REPA alone loses the most of the image-level part.

\emph{3. The gain needs a clean token carrying the image-level part of the image itself, entering before the aligned blocks.}
Removing or weakening any one of these four requirements, with REPA unchanged, pulls \Mloc from 0.399 back toward the 0.325 of REPA alone.
Without the image's own information, \Mloc falls nearly to REPA alone when the trained model's token is zeroed (\cref{fig:supp_probe_t}) and to at most 0.348 with a constant, a class-level, or another same-class image's token (\cref{sec:supp_specificity}). 
With only part of the image-level part, the encoder's [CLS] token (\cref{sec:alt_tokens}), a global summary of the same image, reaches only 0.367; without a clean token, a token noised with the image reaches 0.344.
Without early entry, the gain disappears: the \emph{late token} leaves \Mloc at the level of REPA alone (\cref{tab:interaction,sec:supp_placement}).

\subsection{Finding 3: the source of the FID gain}
\label{sec:supervise}

Turning from the hidden states (Findings~1 and 2) to FID, we find REPA's FID gain mainly in the within-image part.

\paragraph{Setup: aligning one part at a time.}
To tell whether the image-level part or the within-image part improves FID, we train models with REPA's loss restricted to one part, with and without the token (\cref{tab:target}): \emph{image-level REPA} aligns only the image-level part of the projected hidden states, $\bar{\mathbf{h}}$, with $\bar{\mathbf{y}}$, and \emph{within-image REPA} aligns only their within-image parts, $\tilde{\mathbf{h}}^{[n]}$, with $\tilde{\mathbf{y}}^{[n]}$; nothing else of REPA changes.

\paragraph{Observation.} \emph{REPA's FID gain comes mainly from the within-image part (\cref{fig:findings}c).}
Aligning the within-image part alone improves FID far more than aligning the image-level part alone, without the token (over the model with neither, by 20.1 against 9.7) and with it (over the token alone, by 19.9 against 1.4; \cref{tab:interaction}).
The clearest case is image-level REPA with the token: it teaches the image-level part best of all models (\Mglob 0.94) yet barely changes FID, and raising its weight $\lambda$ does not help either (\cref{sec:supp_weight}).

\paragraph{\emph{With the token}, full REPA does at least as well as within-image REPA.}
Against within-image REPA, full REPA with the token reaches a higher \Mloc, a better sFID, and a similar FID (47.03 and 47.55; \cref{tab:interaction}).
Its edge comes from the image-level term: this term makes the blocks use the token more (block~1's attention to the token: 67\% \vs 41\% with within-image REPA; \cref{sec:atten-exp-settings}).

\section{GC-REPA}
\label{sec:method}

Following the rule from Findings~1--3, \emph{supply the global} (Finding~1), \emph{supervise the local} (Findings~2 and~3), \ours (\cref{fig:roles}d) gives REPA the token as an extra input and draws the token at sampling from a token prior. Concretely, the token is the clean patch average of the image itself and enters before block~1, and the unchanged REPA loss supervises the local through its within-image term; with the token, its image-level term does no harm.

\paragraph{Training with the token and REPA.}
\label{sec:training}
\ours minimizes
\begin{equation}
  \mathcal{L}(\theta,\phi) =
  \mathcal{L}_{\text{vel}}(\theta;\mathbf{g})
  +\lambda\,\mathcal{L}_{\text{REPA}}(\theta,\phi;{\mathbf{g}}),
  \label{eq:objective}
\end{equation}
where $\mathbf{g}$ is the token, and $\mathcal{L}_{\text{vel}}(\theta;\mathbf{g})$ and $\mathcal{L}_{\text{REPA}}(\theta,\phi;\mathbf{g})$ are the denoising loss (\cref{eq:vel}) and the REPA loss (\cref{eq:repa}), unchanged, except that $\mathbf{v}_\theta$ now also receives $\mathbf{g}$.
Specifically, the embedding $\mathbf{e}$ of $\mathbf{g}$ (\cref{eq:token}) is prepended to the $N$ patch embeddings $\mathbf{p}_t$ of $\mathbf{z}_t$ and processed by the aligned blocks, $\mathrm{Blocks}_{1:\ell}$, \ie, blocks $1,\dots,\ell$ of $\mathbf{v}_\theta$ applied in sequence (the remaining blocks follow as usual):
$\big[\mathbf{h}_t^{[0]};\,\mathbf{h}_t^{[1]};\,\dots;\,\mathbf{h}_t^{[N]}\big] = \mathrm{Blocks}_{1:\ell}\big([\mathbf{e};\,\mathbf{p}_t]\big)$.
The REPA loss is computed on the hidden states, $\mathbf{h}_t^{[1]},\dots,\mathbf{h}_t^{[N]}$, which now depend on $\mathbf{g}$ through self-attention.

\ours is simple, and what it adds to training is small: the change to REPA's training code is a few lines (\cref{alg:train}), and the only new parameters are the linear layer and the LayerNorm producing $\mathbf{e}$ in \cref{eq:token}: 0.59M, 0.79M, and 0.89M for SiT-B/2, L/2, and XL/2.
The architecture, $\lambda=0.5$ and the setting of the aligned blocks follow REPA.

\paragraph{Sampling with the token prior.}
\label{sec:sampling}\label{sec:prior}
At sampling, there is no image to compute $\mathbf{g}$ from, so we first sample $\mathbf{g}$ (\cref{eq:token}) for the class and then generate the image given $\mathbf{g}$:
\begin{equation}
  p(\mathbf{z}_*\mid c) = \int p_\theta(\mathbf{z}_*\mid c,\mathbf{g})\;p_\psi(\mathbf{g}\mid c)\,\mathrm{d}\mathbf{g},
  \label{eq:factor}
\end{equation}
where $p_\theta$ is the distribution that $\mathbf{v}_\theta$ samples from when it receives $\mathbf{g}$, and the \emph{token prior} $p_\psi$ is a 96M-parameter flow model on $\mathbb{R}^{D}$, trained once on the pairs $(\mathbf{g},c)$ of the training images. 
The drawn $\mathbf{g}$ is renormalized to unit length (\cref{sec:implementation-gc-repa}).
The token prior adds at most 0.2\% to REPA's sampling compute, and since $\mathbf{g}$ depends only on the encoder, one token prior serves every diffusion transformer that uses the same encoder.

\section{Related Work}
\label{sec:related}

We group prior methods by the role of the representation (more in \cref{sec:supp_related}).
\textbf{Representations as alignment targets.}
REPA~\cite{yu2025repa} aligns the hidden states of a diffusion transformer~\cite{peebles2023dit} with the representation of a pretrained encoder.
Later work changes what is aligned or when, \eg, HASTE~\cite{wang2025haste}, REED~\cite{wang2025reed}, sREPA~\cite{xu2026srepa}, and iREPA~\cite{singh2026irepa}; concurrently, LAP~\cite{wang2026lap} removes the target's locally predictable content.
All of these use the representation only as an alignment target. 
Closest to us, iREPA~\cite{singh2026irepa} finds that the representation's spatial structure matters more for REPA's FID than its global semantics, and partly \emph{removes} global components from REPA's target.
Our Finding~3 instead aligns each part of REPA's target separately and finds REPA's gain mainly in the within-image part.
Guided by Findings~1--3, where iREPA removes global components from the target, \ours keeps the image-level part there and also \emph{supplies} it as a condition.
\textbf{Representations as conditions.}
Some methods condition on representations of given images, retrieved from a database~\cite{casanova2021icgan,blattmann2022rdm} or provided for control~\cite{karthikeyan2026repcond}; RCG~\cite{li2024rcg} instead samples them from a learned prior.
Like RCG, \ours samples a clean condition from a separate prior, but pairs it with REPA's alignment target and analyzes how the two roles interact on each part of the representation.
\textbf{Joint generation of representation and image.}
ReDi~\cite{kouzelis2025redi} and REG~\cite{wu2025reg} noise representations with the latent and denoise both.
The representation is thus noisy at high noise, where the aligned blocks most need the image-level part (Findings~1 and 2); \ours's token stays clean and need not be generated jointly.

\section{Experiments}
\label{sec:experiments}

\subsection{Setup}
\label{sec:setup}

We train SiT-B/2, L/2, and XL/2~\cite{ma2024sit} on ImageNet $256\times256$~\cite{deng2009imagenet} (1.28M training images, 1,000 classes) in the latent space of the Stable Diffusion VAE~\cite{rombach2022ldm}.
We follow the evaluation protocol of REPA, HASTE, and REED~\cite{yu2025repa,wang2025haste,wang2025reed}: 50K samples with random labels, the 250-step SDE Euler--Maruyama sampler, and ADM's evaluator~\cite{dhariwal2021adm}. Implementation details are in \cref{sec:supp_impl}.

\paragraph{Baselines.}
We compare with SiT and REPA, with methods that change REPA's target or schedule (HASTE, REED, iREPA, sREPA, SARA, REPA-E) or generate the representation jointly with the image (ReDi); \cref{sec:supp_related} details each.
We report the baselines' numbers from their papers.
\textbf{Metrics.} Following REPA, we report FID~\cite{heusel2017fid} as the main metric, together with sFID~\cite{nash2021sfid}, a spatial variant of FID, Inception Score (IS)~\cite{salimans2016is}, and precision and recall~\cite{kynkaanniemi2019prec}.

\subsection{Comparison with baselines}
\label{sec:main_results}

\begin{table*}[t]
  \centering
  \footnotesize
  \setlength{\tabcolsep}{3.4pt}
  \renewcommand{\arraystretch}{0.92}
  \begin{tabular}{@{}l cccc c cccc c cccc@{}}
    \toprule
    & \multicolumn{4}{c}{SiT-B/2 (130M)} && \multicolumn{4}{c}{SiT-L/2 (458M)} && \multicolumn{4}{c}{SiT-XL/2 (675M)} \\
    \cmidrule{2-5}\cmidrule{7-10}\cmidrule{12-15}
    Method & 50K & 100K & 200K & 400K && 50K & 100K & 200K & 400K && 50K & 100K & 200K & 400K \\
    \midrule
    SiT~\cite{ma2024sit} & -- & -- & -- & 33.0 && -- & -- & -- & 18.8 && -- & -- & -- & 17.2 \\
    + REPA~\cite{yu2025repa} & 78.2 & 49.5 & 33.2 & 24.4 && 55.4 & 24.1 & 14.0 & 10.0 && 52.3 & 19.4 & 11.1 & 7.9 \\
    + HASTE~\cite{wang2025haste} & -- & 39.9 & 25.7 & 19.6 && -- & 19.6 & 12.1 & 8.9 && -- & 15.9 & 9.9 & 7.3 \\
    + REED~\cite{wang2025reed} & 66.0 & 36.8 & 24.6 & 19.5 && 41.7 & 19.1 & 12.1 & 8.9 && 38.0 & 16.0 & 9.8 & 7.3 \\
    + iREPA~\cite{singh2026irepa} & -- & 43.4 & -- & 21.4 && -- & 20.3 & -- & 9.4 && -- & 17.0 & -- & 7.5 \\
    + sREPA~\cite{xu2026srepa} & -- & 39.9 & -- & 20.5 && -- & -- & -- & 8.7 && -- & -- & 9.4 & 7.2 \\
    + REPA-E~\cite{leng2025repae} & -- & 34.8 & -- & -- && -- & 16.3 & -- & -- && -- & 12.8 & 7.2 & 4.1 \\
    + ReDi~\cite{kouzelis2025redi} & -- & -- & -- & -- && -- & -- & -- & 9.4 && 56.1 & 23.1 & 12.6 & 7.5 \\
    + REPA + ReDi~\cite{kouzelis2025redi} & -- & -- & -- & -- && -- & -- & -- & -- && 44.8 & 15.2 & 8.3 & 5.3 \\
    \midrule
    + \ours w/o REPA (token only) & 67.4 & 39.6 & 24.3 & 16.5 && 48.8 & 21.9 & 11.5 & 7.3 && 44.0 & 18.2 & 9.5 & -- \\
    \rowcolor{cvprblue!8}
    + \ours (ours) & \textbf{47.0} & \textbf{23.9} & \textbf{13.4} & \textbf{9.3} && \textbf{21.6} & \textbf{8.5} & \textbf{4.9} & \textbf{3.5} && \textbf{19.3} & \textbf{6.8} & \textbf{4.0} & \textbf{3.0} \\
    \bottomrule
  \end{tabular}
  \caption{\textbf{FID without guidance across training iterations and model sizes}.
  Baselines: published numbers; the parameter counts are those of SiT, and \ours has an extra 96M token prior shared by all sizes; ``--'': not reported.
  }
  \label{tab:main}
\end{table*}

\paragraph{Evaluation across iterations.}
\ours has the lowest FID among the reported results of all three model sizes (\cref{tab:main}):
the margin is largest early (47.0 against 66.0 for REED and 78.2 for REPA at SiT-B/2, 50K) and remains large at 400K (9.3 against 19.5 and 24.4 at SiT-B/2; 3.5 against 8.7 for sREPA and 10.0 for REPA at SiT-L/2).
Our method converges fast: \ours reaches REPA's published 400K FID within 100K at all three sizes, and at SiT-XL/2 its 200K FID (4.0) beats REPA's after 4M iterations (5.9~\cite{yu2025repa}).
Some smaller models with \ours beat larger baselines at 400K, even counting the token prior: SiT-B/2 with \ours (227M in total) beats REPA's SiT-L/2 (458M; 9.3 \vs 10.0), and SiT-L/2 with \ours (555M) beats every SiT-XL/2 baseline (675M; 3.5 \vs 4.1 for the best, REPA-E). Neither the token nor REPA suffices alone: without REPA (``\ours w/o REPA'' in \cref{tab:main}), the token alone has a much worse FID at all reported iterations of all three sizes (16.5 \vs 9.3 at SiT-B/2, 400K), as does REPA alone (24.4).

\begin{table}[t]
  \centering
  \footnotesize
  \setlength{\tabcolsep}{4.3pt}
  \renewcommand{\arraystretch}{0.92}
  \begin{tabular}{@{}lccccc@{}}
    \toprule
    Method & FID$\downarrow$ & sFID$\downarrow$ & IS$\uparrow$ & Prec.$\uparrow$ & Rec.$\uparrow$ \\
    \midrule
    \multicolumn{6}{@{}l}{\emph{SiT-B/2, 400K iterations}} \\
    + REPA~\cite{yu2025repa} & 24.4 & 6.40 & 59.9 & 0.59 & \textbf{0.65} \\
    + HASTE~\cite{wang2025haste} & 19.6 & 6.38 & 73.0 & 0.62 & 0.64 \\
    + REED~\cite{wang2025reed} & 19.5 & -- & 75.9 & 0.61 & \textbf{0.65} \\
    + iREPA~\cite{singh2026irepa} & 21.4 & 6.77 & 70.9 & 0.60 & \textbf{0.65} \\
    + sREPA~\cite{xu2026srepa} & 20.5 & 6.53 & 71.2 & 0.61 & \textbf{0.65} \\
    \rowcolor{cvprblue!8}
    + \ours (ours) & \textbf{9.3} & \textbf{5.52} & \textbf{121.6} & \textbf{0.70} & 0.62 \\
    \midrule
    \multicolumn{6}{@{}l}{\emph{SiT-L/2, 400K iterations}} \\
    + REPA~\cite{yu2025repa} & 10.0 & 5.20 & 109.2 & 0.69 & 0.65 \\
    + HASTE~\cite{wang2025haste} & 8.9 & 5.18 & 118.9 & 0.69 & \textbf{0.66} \\
    + REED~\cite{wang2025reed} & 8.9 & -- & 122.4 & 0.69 & \textbf{0.66} \\
    + iREPA~\cite{singh2026irepa} & 9.4 & 5.11 & 113.9 & 0.69 & \textbf{0.66} \\
    \rowcolor{cvprblue!8}
    + \ours (ours) & \textbf{3.5} & \textbf{4.83} & \textbf{179.3} & \textbf{0.78} & 0.61 \\
    \midrule
    \multicolumn{6}{@{}l}{\emph{SiT-XL/2, 400K iterations}} \\
    + REPA~\cite{yu2025repa} & 7.9 & 5.06 & 122.6 & 0.70 & 0.65 \\
    + HASTE~\cite{wang2025haste} & 7.3 & 5.05 & 128.7 & 0.72 & 0.64 \\
    + REED~\cite{wang2025reed} & 7.3 & -- & 134.5 & 0.70 & \textbf{0.66} \\
    + iREPA~\cite{singh2026irepa} & 7.5 & 4.89 & 128.6 & 0.71 & 0.65 \\
    + REPA-E~\cite{leng2025repae} & 4.1 & \textbf{4.60} & 161.8 & 0.76 & 0.62 \\
    + ReDi~\cite{kouzelis2025redi} & 7.5 & 5.1 & 129.5 & 0.72 & 0.62 \\
    + REPA + ReDi~\cite{kouzelis2025redi} & 5.3 & 4.9 & 149.8 & 0.74 & 0.63 \\
    \rowcolor{cvprblue!8}
    + \ours (ours) & \textbf{3.0} & 4.78 & \textbf{186.2} & \textbf{0.79} & 0.61 \\
    \bottomrule
  \end{tabular}
  \caption{\textbf{All metrics at 400K iterations}, without guidance.}
  \label{tab:full}
\end{table}

\begin{figure}[t]
  \centering
  {\setlength{\parindent}{0pt}
    \includegraphics[width=0.21\linewidth]{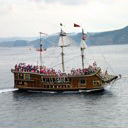}\hspace{1.5pt}\includegraphics[width=0.21\linewidth]{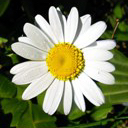}\hspace{1.5pt}\includegraphics[width=0.21\linewidth]{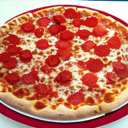}\hspace{1.5pt}\includegraphics[width=0.21\linewidth]{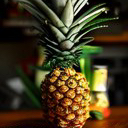}\par
  }
  \caption{\textbf{Samples of SiT-XL/2 with \ours after 400K iterations}. More samples are in \cref{sec:supp_samples}.}
  \label{fig:samples_cmp}
\end{figure}

\paragraph{All metrics at 400K iterations.}
\ours leads in FID, sFID, IS, and precision at all model sizes, with one exception: at SiT-XL/2, REPA-E, which also tunes the VAE, has a lower sFID (\cref{tab:full}).
\ours's recall is 0.01--0.05 lower than the baselines'.
The cause is the token prior: its tokens vary slightly less than those of real images (analysis in \cref{sec:ablations}).
\Cref{fig:samples_cmp} shows samples of SiT-XL/2 with \ours.

\begin{table}[t]
  \centering
  \footnotesize
  \setlength{\tabcolsep}{3.5pt}
  \begin{tabular}{@{}lcccccc@{}}
    \toprule
    & \multicolumn{2}{c}{Parameters (M)} & \multicolumn{2}{c}{Training (iter./s)} & \multicolumn{2}{c}{Sampling (img/s)} \\
    \cmidrule(lr){2-3}\cmidrule(lr){4-5}\cmidrule(l){6-7}
    Model & REPA & \ours & REPA & \ours & REPA & \ours \\
    \midrule
    SiT-B/2 & 130 & 131\,+\,96 & 2.79 & 2.78 & 5.79 & 5.76 \\
    SiT-L/2 & 458 & 459\,+\,96 & 1.16 & 1.16 & 2.11 & 2.10 \\
    SiT-XL/2 & 675 & 676\,+\,96 & 0.82 & 0.82 & 1.47 & 1.47 \\
    \bottomrule
  \end{tabular}
  \caption{\textbf{Cost of REPA and \ours.}
  Parameters: the diffusion transformer, plus the 96M token prior of \ours (trained once in 13 GPU-minutes, shared by all sizes).
  Speeds on one GPU: training at batch size 256; sampling includes the token prior (\cref{sec:supp_compute}).}
  \label{tab:cost}
\end{table}

\paragraph{Cost of the token prior.}
The token prior adds 96M parameters but little compute, since it generates one 768-dimensional vector per image, while the diffusion transformer processes 256 patches with at least 768 dimensions each.
With the token prior and the extra token, \ours trains as fast as REPA and samples at most 0.5\% slower (\cref{tab:cost,sec:supp_compute}).

Even a small token prior helps, and the gain saturates at our default size. A token prior ten times smaller (9.2M parameters) still keeps a large gain over REPA: the same SiT-B/2 with \ours at 400K reaches 16.15 FID, against 24.4 for REPA; the 96M token prior improves it further to 9.32, and a 192M one gives no further gain (9.30).

\begin{table}[t]
  \centering
  \footnotesize
  \setlength{\tabcolsep}{3.6pt}
  \begin{tabular}{@{}lcccccc@{}}
    \toprule
    Method & Iter. & FID$\downarrow$ & sFID$\downarrow$ & IS$\uparrow$ & Prec.$\uparrow$ & Rec.$\uparrow$ \\
    \midrule
    + REPA$^\S$~\cite{yu2025repa} & 400K & 1.98 & 4.60 & 263.0 & 0.80 & 0.61 \\
    + iREPA~\cite{singh2026irepa} & 400K & 1.93 & 4.59 & 268.8 & 0.80 & 0.60 \\
    + sREPA~\cite{xu2026srepa} & 400K & 2.25 & 4.65 & 257.5 & {0.83} & 0.58 \\
    \rowcolor{cvprblue!8}
    + \ours (ours) & 400K & \textbf{1.78} & 4.62 & 245.8 & \textbf{0.84} & \textbf{0.62} \\
    + REPA~\cite{yu2025repa} & 1M & 1.96 & \textbf{4.49} & 264.0 & 0.82 & 0.60 \\
    + REED~\cite{wang2025reed} & 1M & 1.80 & -- & 267.5 & 0.81 & 0.61 \\
    + SARA~\cite{chen2025sara} & 1M & 1.82 & 4.51 & \textbf{279.8} & {0.83} & 0.60 \\
    + sREPA~\cite{xu2026srepa} & 1M & 1.91 & 4.50 & 271.8 & {0.83} & 0.60 \\
    \bottomrule
  \end{tabular}
  \caption{\textbf{SiT-XL/2 with guidance}.
  Guidance: baselines as published. $^\S$Re-run by iREPA~\cite{singh2026irepa}.
  }
  \label{tab:guided}
\end{table}

\paragraph{Evaluation of sampling with classifier-free guidance (CFG).}
\Cref{tab:guided} compares SiT-XL/2 models with CFG~\cite{ho2022cfg}.
\ours uses two guidance scales, one for the class and one for the token, over an interval of noise levels~\cite{kynkaanniemi2024interval} (\cref{sec:supp_cfg}).
At 400K iterations, \ours reaches 1.78 FID, on par with REED (1.80) and SARA (1.82) at 1M iterations.
The gain holds without the two scales: with REPA's guidance setting (one scale, 1.8, for $t\le0.7$~\cite{yu2025repa}), \ours reaches 1.80, still below the guided FIDs of REPA in \cref{tab:guided} (1.98 at 400K, re-run by iREPA, and 1.96 at 1M).

\paragraph{Beyond class labels: text-to-image generation.}
Following REPA's text-to-image experiment, we train MMDiT~\cite{esser2024sd3} from scratch on MS-COCO~\cite{lin2014coco} for 150K iterations (\cref{sec:supp_t2i}).
With the 50-step ODE sampler, \ours reaches FID 4.46, lower than REPA both as REPA reports it and in our reproduction with the same settings, and lower than MMDiT alone (\cref{tab:supp_t2i}).

\subsection{Ablation studies}
\label{sec:ablations}

\paragraph{The performance gain holds with other encoders.}
\label{sec:encoders}
We replace DINOv2-B with a larger encoder, DINOv2-L~\cite{oquab2024dinov2} or CLIP-L~\cite{radford2021clip}, each with its own token prior.
Against REPA alone (\mbox{SiT-B/2}, 400K iterations), \ours lowers FID from 21.88 to 13.04 with DINOv2-L and from 23.54 to 16.03 with \mbox{CLIP-L}, and it also improves sFID, IS, and precision with both encoders. More details are in \cref{sec:supp_encoders}.

\paragraph{With the token, REPA helps denoising more, mostly at high noise.}
With the token, REPA lowers the denoising loss $\mathcal{L}_{\text{vel}}$ more than without the token, by 0.0041 against 0.0017 at SiT-B/2 and by 0.0080 against 0.0065 at SiT-L/2, and the reduction comes from the higher noise levels (\cref{sec:supp_denoise}).
In the aligned blocks, the gradient of the REPA loss also points more in the direction of the gradient of the denoising loss with the token: their mean cosine is 0.285 with the token against 0.197 without it at SiT-B/2, and the increase comes from high noise (\cref{sec:supp_grad}).

\begin{figure}[t]
  \centering
  \begin{tikzpicture}
    \node[inner sep=0pt, anchor=north west] (img) at (0,0) {\includegraphics[width=0.72\linewidth]{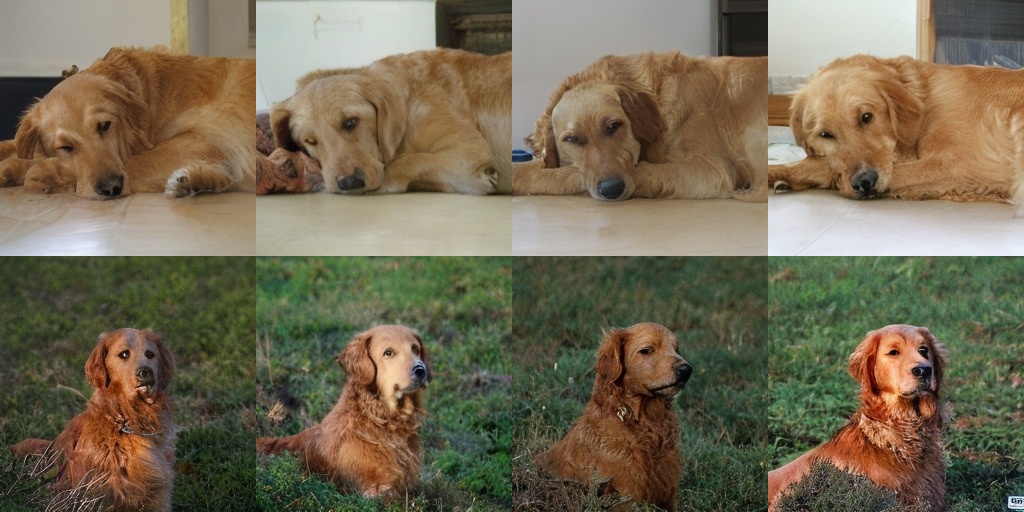}};
    \draw[->, thick] ([xshift=-3pt]img.north west) -- ([xshift=-3pt]img.south west)
      node[midway, rotate=90, above, font=\scriptsize, inner sep=2pt] {varying token};
    \draw[->, thick] ([yshift=3pt]img.north west) -- ([yshift=3pt]img.north east)
      node[midway, above, font=\scriptsize, inner sep=2pt] {varying noise};
  \end{tikzpicture}
  \caption{\textbf{The token sets the scene; the noise sets the details} (golden retriever).
  Each row shares one token from the token prior, each column one initial noise $\mathbf{z}_1$.}
  \label{fig:tokgrid}
\end{figure}

\paragraph{The token sets the scene.}
Within a class, \cref{fig:tokgrid,fig:supp_tokgrid} vary the token and the initial noise $\mathbf{z}_1$ separately, with the deterministic ODE sampler.
We find that changing the token changes the scene (layout and pose); changing the noise changes mainly details, such as textures.
This illustrates that the token carries the image-level part, and that the variety within a class comes mostly from the tokens.

\paragraph{The recall cost comes mainly from the token prior.}
At sampling, the token prior draws the token and the diffusion transformer generates the image from it, so the recall cost could come from either stage; we find that it comes mainly from the token prior.
To see which stage lowers recall, we fix the diffusion transformer, the SiT-B/2 with \ours of \cref{tab:full} (400K iterations), and change only the tokens it receives at sampling (\cref{sec:supp_sources}).
Given tokens of random real images, which spread within a class as real images do, the same diffusion transformer reaches a recall of 0.67, above REPA (0.65, \cref{tab:full}); hence, the diffusion transformer itself does not lower recall.

The recall cost comes instead from the token prior.
With the prior's tokens, the same model's recall is lower than with real ones (0.62 \vs 0.67). This is because these tokens vary less within a class: on average, a token from the prior has a cosine similarity of 0.718 to the mean token of its class, against 0.674 for the token of a real image.
Tokens that vary less give a lower recall; at the extreme, the class-mean token, shared by all samples of a class, gives a recall of only 0.16.

The FID gain does not come from lower diversity, which recall measures: with tokens of random real images, \ours's images are more diverse than those of REPA (recall 0.67 \vs 0.65, \cref{tab:full}), yet its FID stays far better (10.20 \vs 24.4).

\paragraph{No sign of memorization in two checks.}
With tokens of held-out validation images, which the model never saw, SiT-B/2 with \ours at 400K iterations does nearly as well as with tokens of training images (FID 11.66 \vs 10.20; recall 0.64 \vs 0.67), so the model generalizes to tokens it has not seen.
Additionally, a generated image copied from a training image would nearly match its source, yet we find that generated images lie, on average, farther from their nearest training image than real images outside the token prior's training set do (cosine 0.70 \vs 0.80; \cref{sec:supp_mem}).

\section{Conclusion}
\label{sec:conclusion}

In this work, we ask whether a pretrained representation's two roles in a diffusion transformer, as a condition and as REPA's alignment target, conflict when combined.
Splitting the representation into an image-level part, which both roles carry, and a within-image part, which only REPA supervises, we find that the two roles cooperate in teaching both parts, giving the rule \emph{supply the global, supervise the local}.
\ours follows this rule with a clean token of the image's own image-level part and REPA's unchanged loss, drawing the token at sampling from a separate token prior. 
Though simple, \ours improves generation quality and accelerates training convergence over baselines.
\textbf{Limitations and future work.}
\ours needs a separate token prior at sampling, which slightly lowers recall.
We test the rule with one patch-average token and REPA's loss; other conditions and alignment losses are next steps.

{
    \small
    \bibliographystyle{ieeenat_fullname}
    \bibliography{main}
}

\input{appendix_labels}
\end{document}